# Multi-Objective Design Optimization of Non-Pneumatic Passenger Car Tires Using Finite Element Modeling, Machine Learning, and PSO/Bayesian Optimization Algorithms

Priyankkumar Dhrangdhariya[1*], Soumyadipta Maiti[1], Venkataramana Runkana[1]

[1]*Address: TCS Research, Phase 3, Hinjawadi Rajiv Gandhi Infotech Park, Hinjawadi, Pune, Pimpri-Chinchwad, Maharashtra, India-411057*

[*]Corresponding author: priyankkumar.dhrangdhariya@tcs.com

## Abstract

Non-pneumatic tires (NPTs) offer a promising alternative to pneumatic tires; however, their discontinuous spoke structures present challenges in stiffness tuning, durability, and high-speed vibration. This study introduces an integrated generative-design and machine-learning-driven framework to optimize UPTIS-type spoke geometries for passenger vehicles. Upper and lower spoke profiles were parameterized using high-order polynomial representations, enabling the creation of approximately 250 generative designs through PCHIP-based geometric variation. Machine-learning models—KRR for stiffness and XGBoost for durability and vibration—achieved strong predictive accuracy, reducing the reliance on computationally intensive FEM simulations. Optimization using Particle Swarm Optimization and Bayesian Optimization further enabled extensive performance refinement. The resulting designs demonstrate ±53% stiffness tunability, up to ~50% durability improvement, and ~43% reduction in vibration compared to the baseline. PSO provided fast, targeted convergence, while Bayesian Optimization effectively explored multi-objective trade-offs. Overall, the proposed framework enables systematic development of high-performance, next-generation UPTIS spoke structures.

# 1 Introduction

In recent years, the automotive sector has increasingly explored Non-Pneumatic Tires (NPTs) as potential alternatives to traditional pneumatic tires (PTs). Compared to conventional PTs, NPT concepts are considered more environmentally sustainable, technologically modern, and energy-efficient [Sandberg et. al, 2020]. One notable development in this domain is Michelin's UPTIS (Unique Puncture-Proof Tire System), illustrated in Figure 1, which showcases the application of airless tire technology for passenger vehicles. Since UPTIS does not rely on internal air pressure, it fully eliminates puncture-related failures and reduces downtime associated with tire maintenance [Michelin report]. Additionally, studies have reported that NPT systems can offer reduced rolling resistance and improved ride comfort relative to standard pneumatic tires [Ma J. et. al, 2011, Cho J.R. et. al, 2004, Rhyne T.B. and Cron S.M., 2006, M. Veeramurthy et. al, 2011]. The materials adopted in these next-generation designs are largely renewable, helping to minimize waste generation and overall carbon footprint [J. M. Jafferson and H. Sharma, 2021].

Michelin's "Scrapyard Survey" (2013–2015) highlighted that approximately 20% of the 200 million tires discarded annually fail due to irreparable air-loss issues or uneven wear resulting from incorrect inflation pressures. This finding underscores the growing necessity for tire technologies that do not depend on inflation at all. According to Michelin, a significant portion of these discarded tires could be prevented through the adoption of UPTIS and TWEEL technologies [Edward Brell, 2020]. UPTIS represents a promising direction for future airless tire solutions capable of supporting high-speed vehicle applications.

The elastomeric spokes used in non-pneumatic tires (NPTs) are positioned radially and deform locally to form the tire's ground-contact region. Over time, manufacturers such as Michelin, Bridgestone, and Polaris have introduced a wide range of spoke architectures tailored for different operational needs, including military mobility, earthmoving equipment, lunar exploration, stair-climbing systems, and conventional passenger transport. Numerous studies have demonstrated that spoke geometry strongly governs the mechanical behavior and overall performance of NPTs [Dhrangdhariya P. et. Al, 2021; M. Zmuda et. al, 2019]. Existing literature contains several investigations into how geometric parameters influence NPT behavior, particularly for honeycomb-based spoke designs. For instance, Bezgam et al. (2009) examined how variations in honeycomb cell angle, while maintaining constant wall thickness and load-carrying capability, affect deformation patterns, stress fields, and contact pressure distribution in NPTs. Ju et al. further evaluated different honeycomb cell angles by studying fatigue resistance and localized stress concentrations under vertical loading through FEM simulations. X. Jin et al. (2018) explored both static and dynamic characteristics of three honeycomb-structured NPT configurations, each sharing identical wall thickness or load capacity but differing in geometric detail. Additionally, Ingrole A. et al. (2017) compared multiple structural layouts, including conventional honeycomb, re-entrant auxetic honeycomb, locally reinforced auxetic-strut designs, and hybrid combinations integrating honeycomb and auxetic-strut features. Their results showed that the hybrid configuration offered substantial improvements, achieving nearly three times the compressive strength of a traditional honeycomb sandwich and approximately 65% greater strength than a standalone auxetic structure.

A substantial body of research has focused on optimizing NPT spoke designs through various computational and statistical techniques, including Design of Experiments (DOE), sensitivity-based studies, Response Surface Modeling (RSM), and more recently, artificial intelligence and machine-learning-driven approaches [M. Veeramurthy et. al, 2014, K. Kim et. al, 2015, I. G. Jang et al., B. M. Kwak et. al, 2009, Yang et. al, 2014, A. S. Pramono et. al, 2019]. Across these studies, parameters such as shear-band thickness, spoke thickness, and honeycomb cell angle have commonly been treated as key design variables, often accompanied by volumetric or structural constraints. I. G. Jang et al. (2012) and B. M. Kwak et. al (2009) have specifically advocated the use of structural topology optimization to refine NPT spoke patterns so that the resulting static stiffness closely matches that of conventional pneumatic tires. Their optimization frameworks explored combinations of section numbers, material volume fractions, and weighting factors within nested optimization loops to systematically identify spoke configurations that satisfy the imposed volumetric constraints while achieving the desired mechanical response.

Yang et al. (2004) developed an Artificial Neural Network (ANN) model trained using finite element analysis (FEA) data generated from multiple tire-design configurations. The advantage of employing machine-learning-based prediction in tire development is that it enables rapid and cost-effective exploration of large parametric spaces, allowing designers to evaluate and refine new concepts without performing exhaustive FEM simulations for every design iteration. Similarly, A. S. Pramono et al. (2019) proposed an AI-assisted methodology for airless tire design using three key input parameters—spoke thickness, rhombic angle, and rubber material. In their study, tire deflection and overall stress response were obtained through FEM, after which an ANN was used to learn the nonlinear mapping between inputs and outputs. The trained model was subsequently integrated with a genetic algorithm to identify optimal combinations of parameters for enhanced tire performance.

In our earlier research, we examined how the nonlinear behaviour of polyurethane (PU) materials influences the performance of three distinct spoke architectures—Tweel, Honeycomb, and UPTIS by evaluating ride comfort, stiffness, and damage-related metrics. Among these designs, the Michelin UPTIS exhibited noticeably lower stiffness and greater cushioning capability compared to both the Honeycomb and Tweel configurations, reinforcing that spoke geometry plays a decisive role in determining overall tire performance. Additionally, targeted, filter-based topological optimization of stiffness and durability characteristics had been conducted in prior work [Dhrangdhariya et. al, 2021 & 2023]

Given that the discontinuous nature of NPT spokes inherently amplifies vibration at higher operating speeds, reducing this vibration becomes a crucial factor in any optimization framework. To the best of the authors' knowledge, no published studies have yet addressed comprehensive, multi-feature optimization of UPTIS spoke designs using advanced techniques such as Bayesian Optimization or Particle Swarm Optimization, highlighting a significant research gap in the current literature.

In this study, the geometric profiles of the UPTIS spokes, specifically the upper and lower curve boundaries, were extracted and reformulated using polynomial curve-fitting techniques to obtain precise analytical representations. These polynomial functions were then employed to generate nearly 250 novel spoke configurations through a generative design framework that utilized Piecewise Cubic Hermite Interpolating Polynomial (PCHIP) interpolation for smooth and controlled geometric variation. Machine Learning (ML) models were subsequently developed to evaluate the stiffness and damage characteristics of these designs, trained using finite element (FEM) simulation data from 200 samples, while an additional set of 50 designs was reserved exclusively for testing and verification. Following this, comprehensive topological optimization was carried out using both Bayesian Optimization and Particle Swarm Optimization methodologies. The optimization process incorporated single and multi-objective schemes, such as minimization, maximization, and target-oriented searches, to identify spoke geometries capable of simultaneously delivering the required stiffness, enhanced durability, and reduced vibration response.

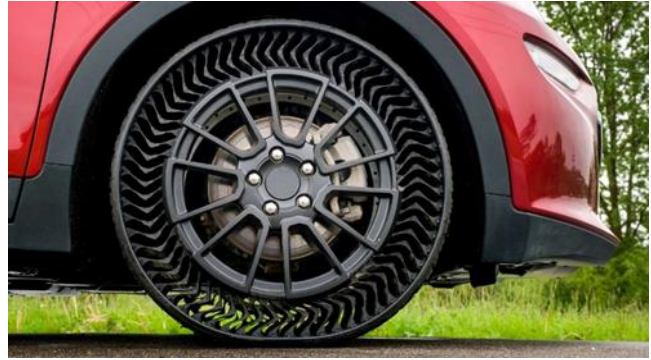

Figure 1. Michelin's NPT-UPTIS mounted on Bolt electric car by GM

## 2 Modeling and simulation of NPT-UPTIS Spoke:

### 2.1 FEM Simulation of NPT-UPTIS Spoke

#### *2.1.1 Computer Aided Design (CAD) of NPT-UPTIS Spoke:*

Non-pneumatic tires (NPTs) typically comprise a hub, shear band, inner and outer reinforcement rings, tread, and a compliant spoke structure that connects the hub to the shear band [Dhrangdhariya et. al, 2021 & 2023] For the present study, the geometric parameters and reference coordinate data of the NPT-UPTIS spoke were extracted from publicly available Michelin UPTIS documentation and digitized as shown in Figure 2. [Nikolov et. al, 2023] The selected tire corresponds to a 215/60 R15 passenger car tire, which specifies a 15-inch rim diameter and an overall tire outer diameter (OD) of approximately 640 mm, with a nominal tread width of 215 mm. Based on these dimensions, the effective spoke length was estimated to be approximately 108 mm. The detailed spoke curvature and coordinate profiles were reconstructed using the Plot Digitizer tool, enabling precise extraction of the spline geometry from Michelin's published illustrations.

Using the digitized coordinate data, the CAD profile of the NPT-UPTIS spoke was generated in ANSYS Design Modeler. The reconstructed profile accurately captures the characteristic curvature, thickness variation, and connection interfaces of the UPTIS spoke architecture, ensuring geometric fidelity for subsequent finite-element modelling and performance evaluation. The final CAD representation used in the analysis is presented in Figure 3.

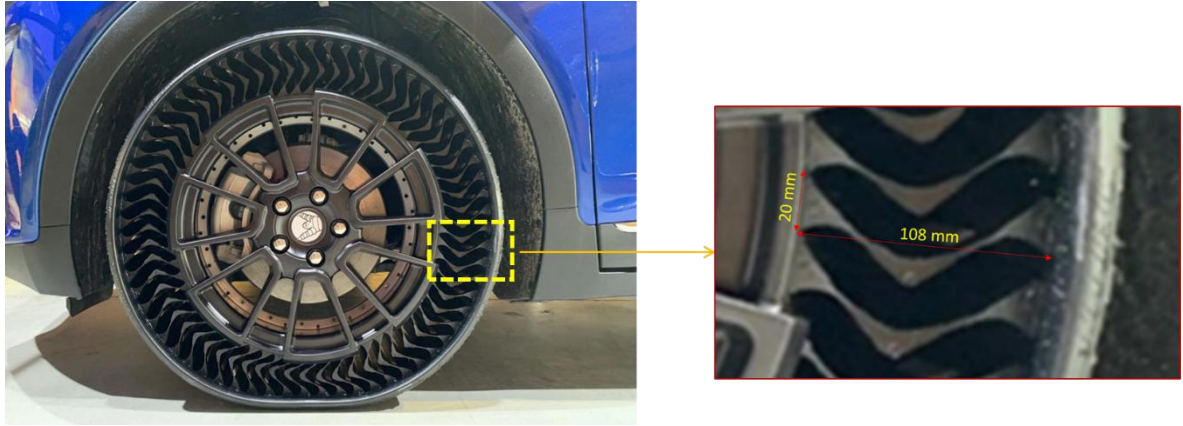

**Figure 2. (a) CAD model of NPT-UPTIS (215/60 R15) (b) Spoke schematic (All dimensions are in mm)**

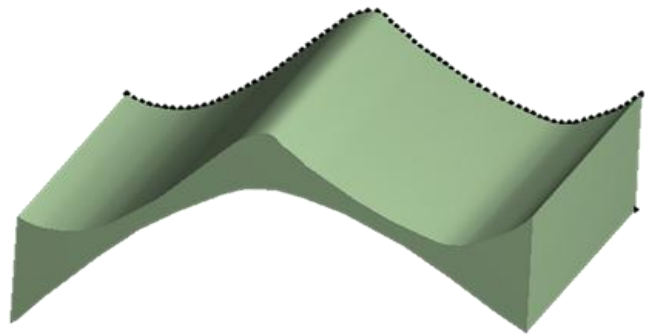

**Figure 3. CAD model of NPT-UPTIS Spoke design**

### 2.1.2 *Material property and hyperelastic curve fitting for NPT Spoke:*

DBDI-based polyurethane (4,4′-dibenzyl diisocyanate) was selected as the spoke material owing to its superior damage resistance and favourable stiffness characteristics, as demonstrated in our earlier studies [Dhrangdhariya et. al, 2021]. Comprehensive material data and constitutive parameters for DBDI polyurethane are available in prior literature Dhrangdhariya et. al, 2021]. Several hyper-elastic constitutive models, such as the Mooney–Rivlin (2, 3, and 5-parameter) formulations, the Yeoh model, Arruda–Boyce model, and the Neo-Hookean model, were previously evaluated for this material [Dhrangdhariya et. al, 2021]. Among these, the Mooney–Rivlin 5-parameter model provided the closest representation of the experimentally measured non-linear stress–strain response of DBDI polyurethane. The corresponding material constants used in this study are summarized in Table 1. As illustrated in Figure 4, the 5-parameter Mooney–Rivlin model offers excellent agreement with the experimental curve, capturing both the initial compliance and the pronounced non-linear stiffening at higher strains.

**Table 1 Mooney Rivlin 5 Parameters material constants for DBDI experimental data**

| | | Hyper-elastic constants | | | | |
|---|---|---|---|---|---|---|
| *Component* | *Model* | *C10* | *C01* | *C20* | *C11* | *C02* |
| **Shear Beam & spokes** | *Mooney Rivlin 5* | -16.565 | 30.572 | -0.0281 | 0.3005 | 3.7316 |

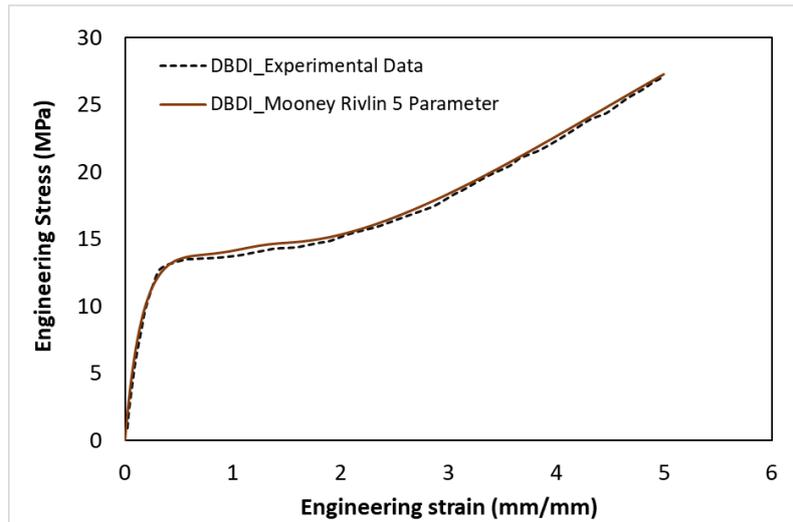

**Figure 4. Stiffness comparison of 2D plane strain and 3D simulation for spoke**

### 2.1.3  *Stiffness and Durability Analysis:*

There are three types of FEM simulations performed in this study, for stiffness and durability analysis 2D Plane Strain simulations were performed in ANSYS Workbench structural mechanics module, while transient analysis was used for Vibration analysis of 3D spokes.

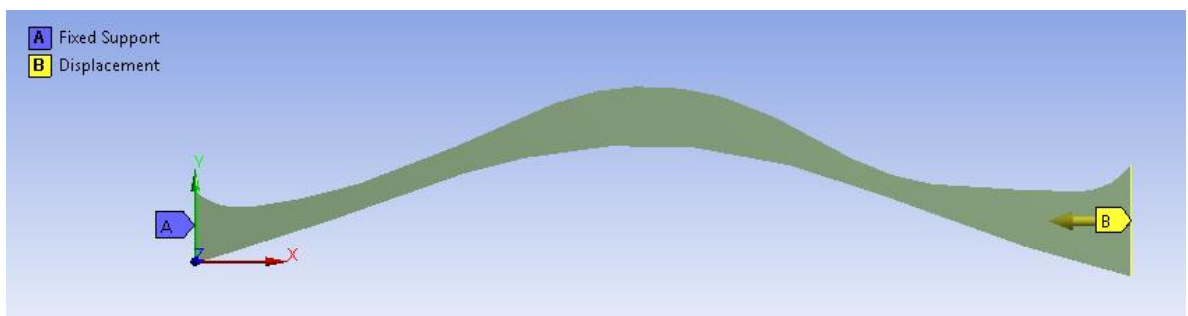

**Figure 5. 2D FEM spoke geometry with boundary conditions**

The 2D CAD profile of NPT-UPTIS spoke have been imported to ANSYS Workbench 16.0. As the spoke geometry is symmetric in the 3$^{rd}$ dimension and the strain in the third dimension is negligible, plane strain condition was assumed and 2D FEM simulations were performed. Left side face of the spoke is restrained with fixed support boundary condition to restrict the translational and rotational movements which is depicted with 'A' and displacement has been applied at the right end face as mentioned with 'B' in Figure 5. The spoke performance has been analysed with both tensile and compressive simulations. In the first case, 50 mm of tensile displacement (50 mm in +X direction) is applied and in the second case 50 mm of compressive displacement (50 mm in -X direction) is applied at the right end side of spoke. The element used for the 2D plane strain FEM simulation is PLANE 182 with mesh size of around 1 mm.

To verify the plane strain assumption, FEM simulation with similar boundary conditions have been carried out with 3D spoke having same 2D profile. The element is used for 3D simulation is SOLID 186 with mesh size of around 5 mm. The stiffness comparison of 2D plane strain assumption and 3D simulation is shown in the Figure 6. The trend of resulting reaction force vs displacement in 2D and 3D simulation is quite matching and hence this plane-strain condition assumption is considered acceptable for all future spoke related FEM simulations.

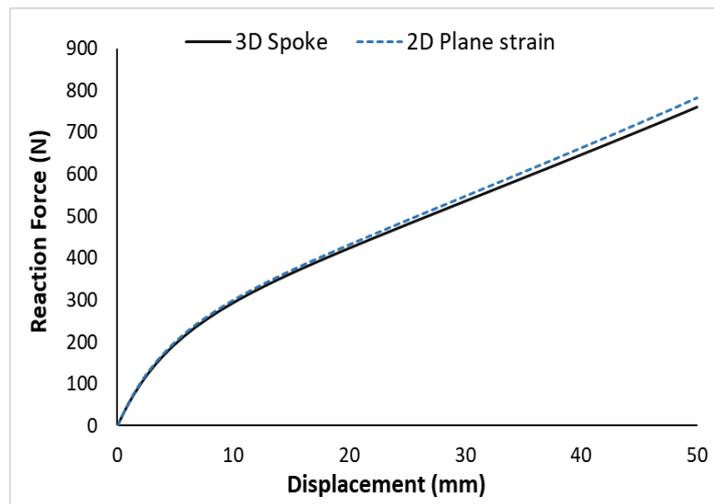

Figure 6. Stiffness comparison of 2D plane strain and 3D simulation for spoke

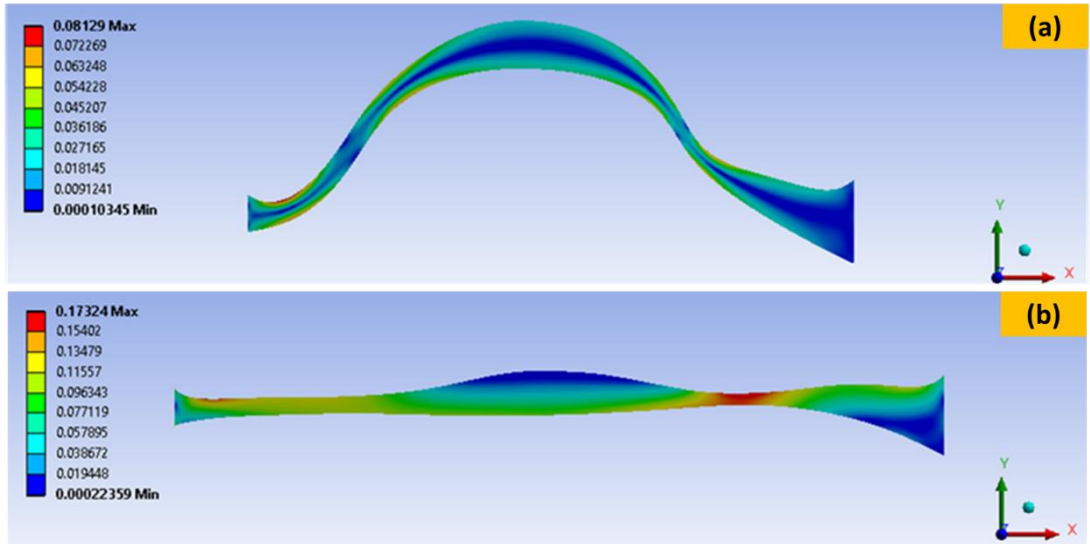

Figure 7. FEM Simulation results of Principal strain variation at (a) 20 mm of displacement in compression & (b) at 20 mm of displacement in Tension

Table 2 Output features from 2D FEM simulation of spoke for ML models

| Property type | Different Output Features | Condition |
| --- | --- | --- |
| Stiffness parameter | Reaction force in compression & tension (RFC & RFT) | At 20 mm of displacement |
| Damage parameter | Strain energy density in compression & tension (SEDC & SEDT) | At 500 N of RFC & At 10000 N of RFT |
| Vibration Parameter | Root Mean Square (RMS) value for FFT Amplitudes | 100 to 470 Hz (Frequency) |

The principal strain variation of 2D spoke in tension and compression at 20 mm of displacement has been captured in the Figure 7. Similarly, a total of 10 different outputs from these 2D FEM simulations as shown in Table 2, have been recorded as output features of the FEM simulations on UPTIS spokes. These different output features were used to train the machine learning (ML) models as explained in Section 4. Reaction force in tension (RFT) and compression (RFC) represent the stiffness property of the spokes. While for predicting the damage performance properties, maximum value of principal stress and strain, octahedral shear strain and strain energy density (SED) have been captured as shown in Table 2. For vibration analysis, RMS of FFT amplitudes were considered as output feature. The performance and usefulness of these damage predicting parameters under the road-displacement analysis for

NPT-UPTIS, NPT-Honeycomb and NPT-TWEEL was briefly explained in the previous work. [Dhrangdhariya et. al 2021]

### 2.1.4 Vibration Analysis of 3D Spoke Design

(a)

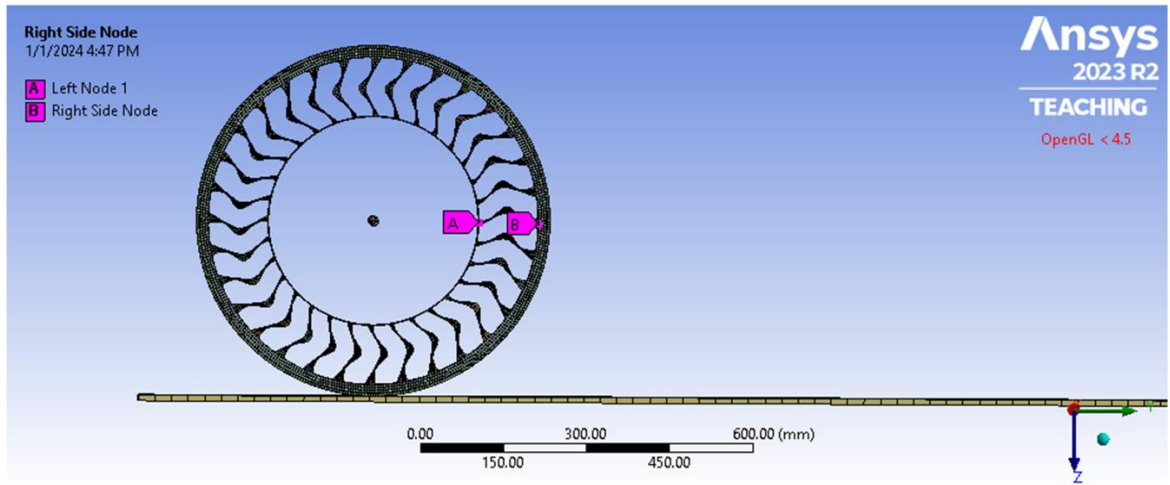

(b)

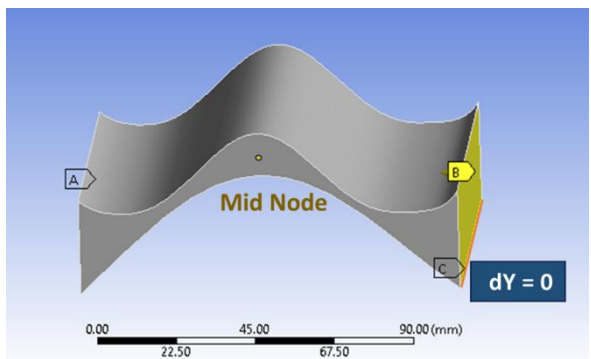

(c)

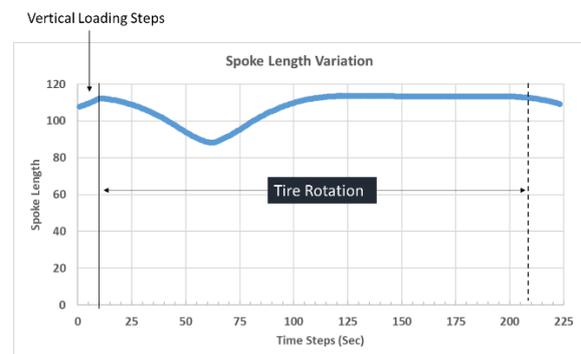

**Figure 8. (a) FEM setup for rolling analysis used to extract spoke displacement during tire rotation, (b) boundary conditions applied in the vibration analysis, and (c) displacement pattern imposed on the spoke to simulate one complete rotation cycle.**

The displacement pattern was extracted using the A and B nodes indicated in the Figure 8 (a). As the tire undergoes rolling, each spoke experiences both expansion and compression. The variation in spoke length, captured between nodes A and B, was used as the boundary condition at the spoke end, illustrated by point B in the Figure 8 (b). The opposite end of the spoke was fixed, as shown at point A, and the vertical displacement along the edge was constrained as indicated at point C.

A transient structural simulation was then performed in ANSYS Workbench for 20 rotational cycles. Two different time durations were considered to represent low-speed and high-speed operating conditions. For the slow-speed case, 20 cycles were completed in 80 seconds. Given that one tire revolution covers approximately 2000 mm, this corresponds to a speed of about 2 km/h. For the high-speed case, the same 20 cycles were simulated in 1 second, representing a speed of roughly 150 km/h.

For both simulations, the displacement response was recorded at the mid-node location shown in the Figure 8 (c).

(a)

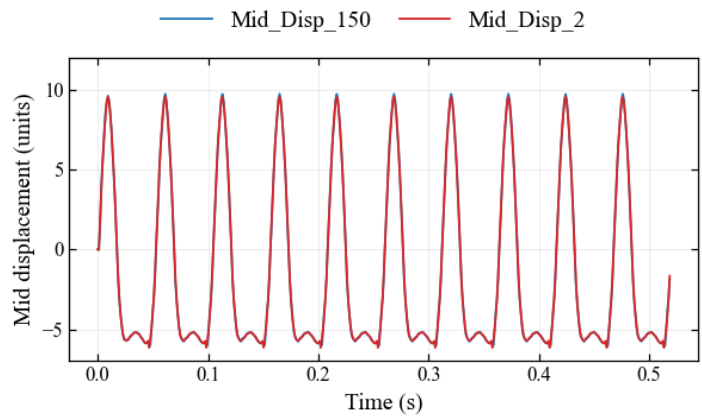

b)

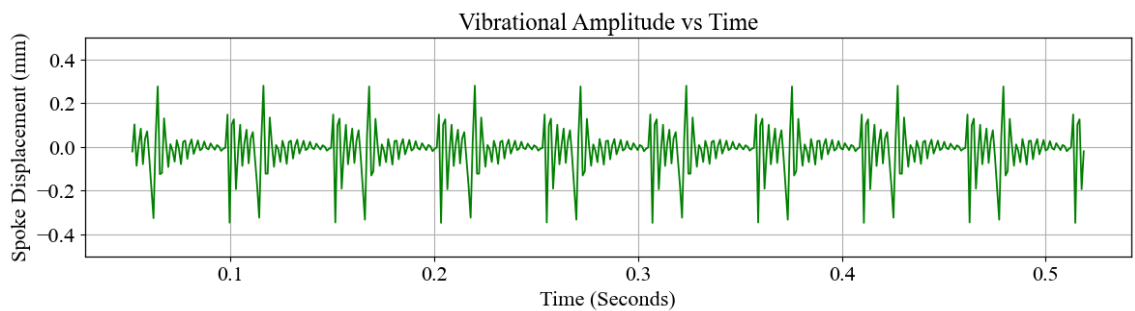

c)

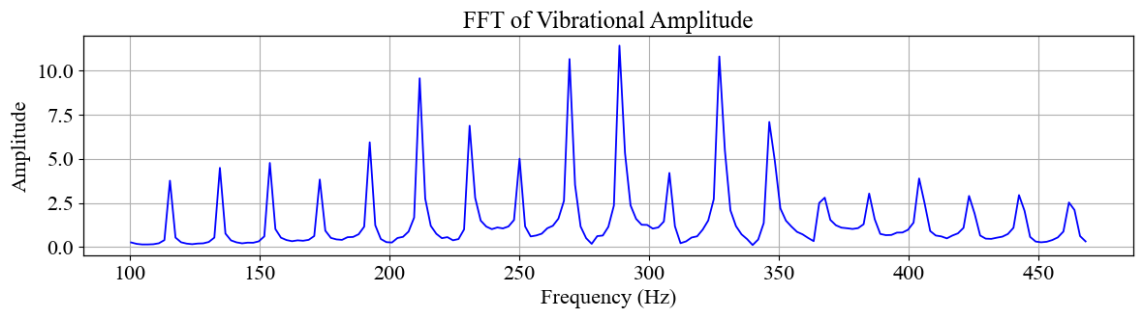

**Figure 9. (a) Mid node displacement pattern for 10 loading cycles at 2 Km/hr and 150 Km/hr, (b)Vibration Amplitude of spoke at mid node at 150 Km/Hr, and (c) FFT Plot of Vibration Amplitude**

The mid-node displacement responses at 150 km/h and 2 km/h were extracted, as shown in Figure 9(a). The difference between these two displacement profiles was computed to obtain the vibration amplitude, presented in Figure 9(b). A Fast Fourier Transform (FFT) was then performed on this vibration amplitude signal, and the corresponding frequency spectrum is shown in Figure 9(c). The resulting RMS value of the FFT amplitudes, calculated over the 100–470 Hz range, is approximately 2.4216, indicating the dominant vibrational content of the Base Design. This value serves as the reference benchmark for assessing improvements achieved through optimized spoke designs. FFT-based vibration characterization has been widely employed by several researchers as well. This RMS value for FFT vibrational amplitude for 100-470 Hz frequency was used as output feature as shown in Table 2. [S. Bezgam et. al, 2009, Rutherford et. al, 2010, Narsimhan A. et. al, 2011, C Lee et. al, 2012]

## 3  Generative Designs for NPT-UPTIS spoke:

### 3.1  Generative Design Approach:

Generative design approach was implemented to create various designs having alteration in the top and bottom spoke curve profiles. Hence, it was required to capture the curvature of top and bottom spoke surfaces with analytical equations.

#### 3.1.1  *Spoke Coordinate extraction:*

The schematic of UPTIS initial reference design with thickness values indicated at different interval is shown in Figure 2(b). To trace the profile of the top and bottom curve of this spoke, around 150 coordinate points of top and bottom curve have been extracted using ANSYS Design Modeler, as shown in Figure 10.

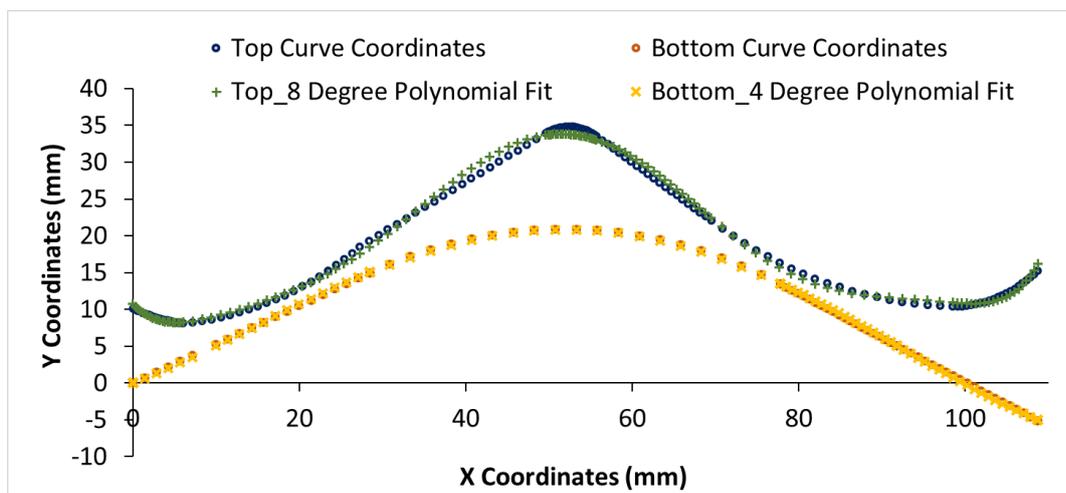

**Figure 10. Top and Bottom curve coordinates with best polynomial curve fitting of UPTIS spoke**

### 3.1.2 *Polynomial curve fitting:*

Figure 10 shows that fitted polynomial equations of 8th order and 4th order are giving good fitting for top and bottom spoke surface coordinates, respectively. The polynomial equations for the top and bottom curve fittings are described in Equations 1 & 2. The coefficients for 8-degree polynomial and 4-degree polynomial are shown in the Table 3.

$$Y = a_0 + a_1 X_t + a_2 X_t^2 + a_3 X_t^3 + a_4 X_t^4 + a_5 X_t^5 + a_6 X_t^6 + a_7 X_t^7 + a_8 X_t^8 \quad (1)$$

$$Y = b_0 + b_1 X_b + b_2 X_b^2 + b_3 X_b^3 + b_4 X_b^4 \quad (2)$$

**Table 3 Coefficients of Polynomial equation for top and bottom curve fit**

| $a_0$ | $a_1$ | $a_2$ | $a_3$ | $a_4$ | $a_5$ | $a_6$ | $a_7$ | $a_8$ |
|---|---|---|---|---|---|---|---|---|
| 10.731 | -0.8865 | 0.10641 | -0.0054192 | 0.00015706 | -2.4846E-06 | 2.1164E-08 | -9.1524E-11 | 1.5803E-13 |

| $b_0$ | $b_1$ | $b_2$ | $b_3$ | $b_4$ |
|---|---|---|---|---|
| 0.029891 | 0.31518 | 0.0052969 | -0.00010414 | 3.5481E-07 |

### 3.1.3 *Generative algorithm for top and bottom spoke profile:*

To generate various spoke profiles with varying top and bottom curves from the polynomial equation, Piecewise Cubic Hermite Interpolating Polynomial (PCHIP) has been utilized. There are 5 interpolation points those have been selected at distance of 0, 27, 54, 81 and 108 mm along X coordinates. The Y coordinates have been assigned with uniform random number generation in the range of -4.0 to +4.0 mm for the top curve and -2.0 to +2.0 mm for the bottom curve. The interpolation points of top and bottom points are shown in the Figure 11 with filled circular marks. PCHIP interpolating polynomial is used to fit these interpolation points. 150 equidistant datapoints have been generated along the X axis of the spoke profiles with the help of PCHIP interpolating spline fitting through the 5 previously specified interpolation points (Figure 11). These coordinates of interpolating polynomial have been added to the initial top and bottom curve coordinates as shown in Equation 3.

$$Y_{Gen} = Y_{initial} + Y_{chip} \quad (3)$$

where,

$Y_{initial}$ = initial reference spoke design Y coordinates of top or bottom spoke surface ($Y_{top}$ or $Y_{bot}$)

$Y_{chip}$ = Y coordinates of interpolating polynomial curve fitted to the 5 interpolation points

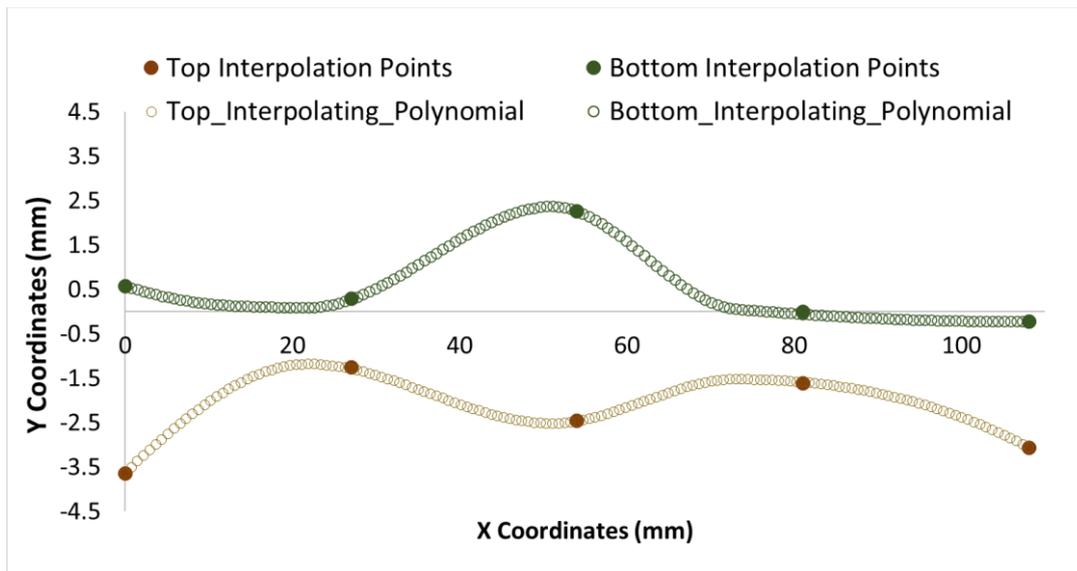

Figure 11 Interpolating polynomial curve fitted to top and bottom interpolation points

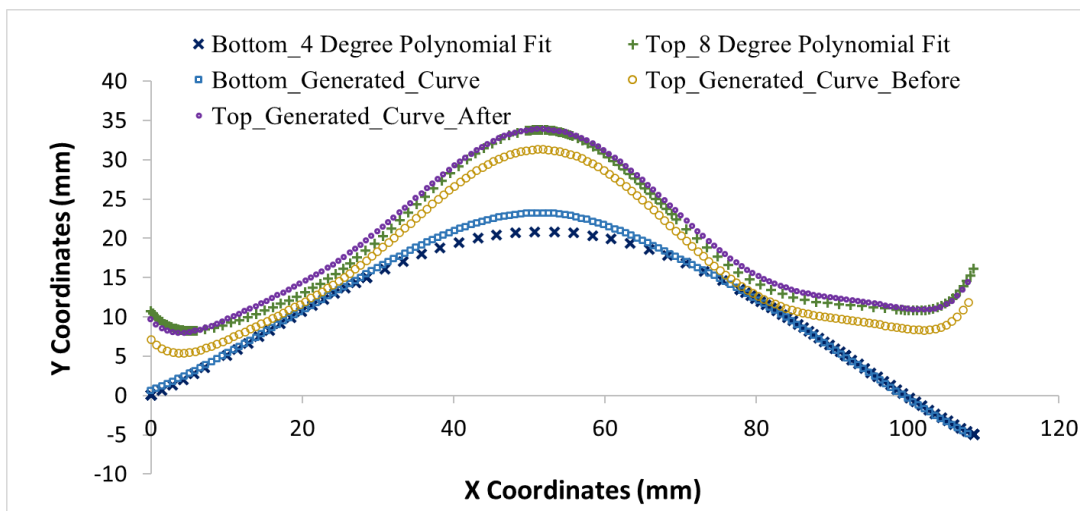

Figure 12 Top and bottom curve of generated spoke profile

In Figure 12, one generated spoke profile after adding the interpolating polynomials to the base-design of top and bottom 8-degree and 4-degree polynomials has been illustrated. To maintain the volume equivalency in all the generated spoke designs, a constraint condition for the area of the new design has been applied such that, the area difference of the generated profile and the initial reference design would not exceed beyond 0.1%. The top and bottom profile curves which are generated from Equation 3 will not necessarily satisfy the area difference convergence criterion. Hence increment or decrement of Y-coordinate values of all points of the top spoke curve in steps of 0.001 mm has been done in each iteration. The increment or decrement of Y coordinates are continued iteratively in a loop until the new generated spoke would attain an area difference within the 0.1% limit, with respect to the initial reference design. In Figure 12, one example of generated curve profile from initial design has been represented. The profile of top spoke surface, which are generated from Equation 3 is termed '*Top_Generated_Curve_before*'. After modifying the dimensions of Y-coordinates of the generated profile for the required area convergence criterion, the final spoke profile '*Top_Generated_Curve_After*' is generated.

With this approach 250 such new spoke designs have been generated. Out of these, 200 randomly created designs have been used to analyse the spoke performance using Finite Element Method (FEM) simulation as explained in the section 2.3.1 and later on to train the Machine Learning (ML) models for mechanical property predictions. The results of FEM have been used as output features for machine learning model developments. 50 designs have been used to test the ML models.

### 3.2 Machine learning models for spoke performance prediction

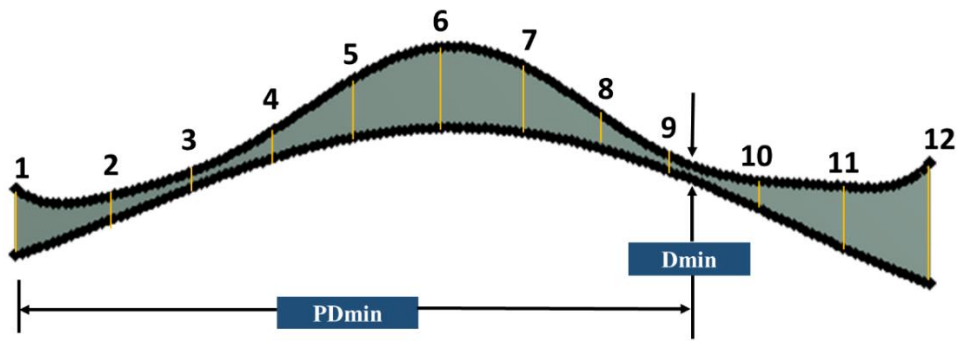

**Figure 13 Machine learning model input features**

*3.2.1 Input/output features:*

Various spoke geometry-based input feature combinations have been tried to model the output features mentioned in Table 2. The combination of 19 input features including 5 bottom interpolation points (explained in section 4.1.3), 12 thickness values at equidistant locations of the spoke, the minimum thickness of spoke ($D_{min}$) and the position of the minimum thickness ($PD_{min}$) as shown in Figure 13, have been found as the best input feature combination compared to other trials. Total 250 designs were used to develop the ML models, where 200 designs were assigned for the train dataset and 50 designs were assigned as the test dataset.

### *3.2.2 Various ML models of regression:*

Various ML Regression models like *Ridge, Lasso, Elastic-Net & Kernel Ridge Regression (KRR), AdaBoost, XgBoost, SVR, GBR, Random Forest, and Desicsion Tree,)* were implemented from scikit learn library using Python. Standardization scaling is used to normalize the input features. GridsearchCV and Kfold cross validation hyper tuning technique have been implemented to find out the optimized and best combination of various ML model parameters which also helps to overcome the overfitting issues.

## 3.3 Optimization Algorithms:

### *3.3.1 Particle Swarm Optimization (PSO)*

Particle Swarm Optimization (PSO) is a population-based stochastic search algorithm inspired by collective swarm behavior, where each particle represents a candidate solution in the design space. The swarm evolves iteratively by updating each particle's velocity and position using a combination of (1) inertia from its previous motion, (2) attraction toward its own historically best position (cognitive learning), and (3) attraction toward the best position found by the entire swarm (social learning). A commonly used PSO formulation is given by the velocity and position update rules:

$$v_i^{t+1} = \omega v_i^t + c_1 r_1 \left(p_i^{best} - x_i^t\right) + c_2 r_2 (g^{best} - x_i^t) \tag{4}$$

$$x_i^{t+1} = x_i^t + v_i^{t+1} \tag{5}$$

Where $x_i^t$ and $v_i^t$ denote the position and velocity of particle $i$ at iteration $t$, $\omega$ is the inertia weight controlling the exploration–exploitation balance, $c_1$ and $c_2$ are the cognitive and social acceleration coefficients, $r_1$ and $r_2$ are random variables that introduce stochasticity. The terms $p_i^{best}$ and $g^{best}$ denote the best solution previously found by particle $i$ and by the swarm, respectively. This mechanism enables efficient global search on complex, non-convex objective landscapes without requiring gradients, which is particularly beneficial for simulation-based design optimization. [Kenedy et. al, 1995]

### *3.3.2 Bayesian Optimization (BO)*

Bayesian Optimization (BO) is a sequential, data-efficient optimization framework designed for expensive black-box objective functions commonly encountered in computational modeling and simulation. BO constructs a probabilistic surrogate model (often a Gaussian Process, GP) from previously evaluated samples, providing a predictive mean $\mu(x)$ and uncertainty $\sigma(x)$ for any candidate $x$. The next evaluation point is selected by maximizing an acquisition function that explicitly balances exploitation (sampling where the surrogate predicts good performance) and exploration (sampling where predictive uncertainty is high). A widely used Acquisition function is Expected Improvement (EI) for single objective optimization and Expected Hypervolume Improvement (EHVI) for multi-objective optimization. Expected Improvement for minimization is defined as follows.

$$EI(x) = (f^* - \mu(x))\Phi\left[\frac{f^* - \mu(x)}{\sigma(x)}\right] + \sigma(x)\phi\left[\frac{f^* - \mu(x)}{\sigma(x)}\right] \qquad (6)$$

where $f^*$ is the best (minimum) observed objective value so far, and $\Phi(\cdot)$ and $\phi(\cdot)$ are the standard normal cumulative distribution and probability density functions, respectively. Intuitively, EI becomes large either when the predicted mean $\mu(x)$ is better than the current best ($f^*$) (exploitation) or when the uncertainty $\sigma(x)$ is large (exploration). This property enables BO to identify near-optimal designs with relatively few expensive simulation evaluations. [Snoek J. et. al, 2012]

## 4 Results & Discussion:

In this section, the performance of ML models, analysis of generative design approach, selection of the optimized spoke designs and their properties have been described in detail.

### 4.1 Performance of ML models

The performance of the ML models described in Section 3.2 was evaluated using the coefficient of determination ($R^2$), as summarized in Figure 14 and Table 4, and the results clearly indicate that the Kernel Ridge Regression (KRR) model provides the highest predictive accuracy for Reaction Force in Compression (RFC) and Reaction Force in Tension (RFT), while in case of durability output parameters like strain energy density in compression (SEDC), strain energy density in tension (SEDT), and the vibration response, XGBoost models demonstrate superior performance. Hence, KRR and XGBoost is selected as the most suitable ML models for the present study based on their overall comparative effectiveness.

To enhance robustness and reduce overfitting, the hyperparameters of both models were optimized using K-Fold cross-validation in conjunction with GridSearchCV. The final optimized hyperparameters for KRR and XGBoost are reported in Table 5.

The predictive accuracy of the selected models is illustrated through parity plots in Figure 15. For RFC and RFT, the predicted and actual values align closely along the 45° reference line, indicating near-ideal agreement, with fitted slopes approaching unity. For SEDC and SEDT, the models also exhibit strong predictive performance, achieving R² values of approximately 0.95 and 0.98, respectively, with correspondingly well-aligned test-set data in the parity plots. In the case of the vibration response, the XGBoost model yields an R² value of about 0.86, which reflects a reasonably strong predictive capability and is consistent with the visual trend observed in the parity plot.

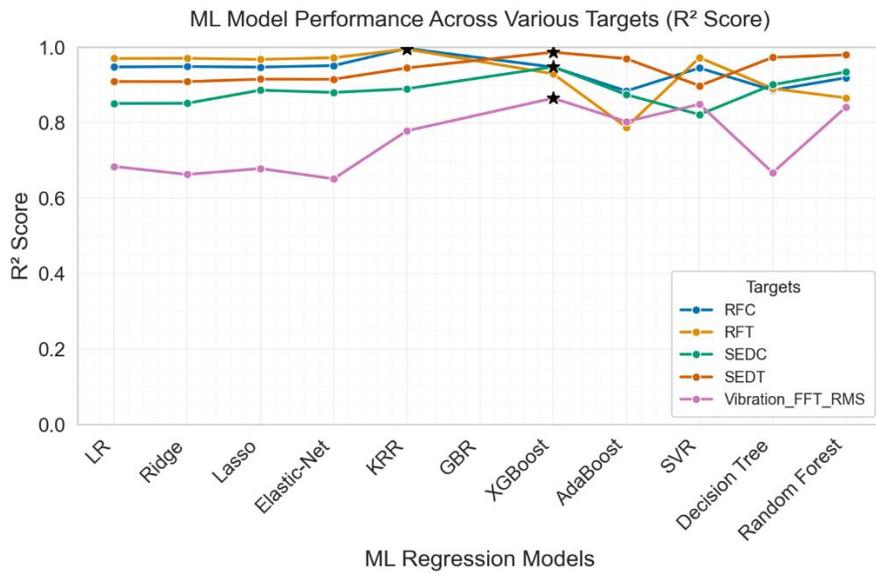

Figure 14 R2 score of all output features for various ML models

Table 4 R2 score for all ouput features for different ML models

| R2 Score | RFC | RFT | SEDC | SEDT | Vibration_FFT_RMS |
|---|---|---|---|---|---|
| **LR** | 0.9479 | 0.9703 | 0.8511 | 0.9092 | 0.6841 |
| **Ridge** | 0.9489 | 0.9708 | 0.8515 | 0.9088 | 0.6628 |
| **Lasso** | 0.9474 | 0.9678 | 0.8863 | 0.9155 | 0.6786 |
| **Elastic-Net** | 0.9514 | 0.9723 | 0.8803 | 0.9152 | 0.6513 |
| **KRR** | **0.997** | **0.995** | 0.8899 | 0.9451 | 0.7784 |
| **XgBoost** | 0.9474 | 0.9297 | **0.948** | **0.987** | **0.865** |
| **Adaboost** | 0.884 | 0.7882 | 0.8744 | 0.9698 | 0.803 |
| **SVR** | 0.9451 | 0.9726 | 0.8216 | 0.898 | 0.8489 |
| **Decision Tree** | 0.8861 | 0.8906 | 0.9008 | 0.9732 | 0.6676 |
| **Random Forest** | 0.9191 | 0.8658 | 0.9346 | 0.98 | 0.8413 |

**Table 5 Optimized hyperparameter of selected ML models for different output features**

| Output Feature | Selected ML model with optimized hyperparameters |
|---|---|
| **RFT** | KRR (kernel: 'poly', $\alpha = 0.001$, $\gamma = 0.01$, degree = 3) |
| **RFC** | KRR (kernel: 'poly', $\alpha = 0.001$, $\gamma = 0.01$, degree = 2) |
| **SEDT** | XgBoost ($\alpha = 0.2$, n_estimator = 150, max_depth = 2) |
| **SEDC** | XgBoost ($\alpha = 0.05$, n_estimator = 100, max_depth = 2) |
| **Vibration_RMS_FFT** | XgBoost ($\alpha = 0.05$, n_estimator = 150, max_depth = 2) |

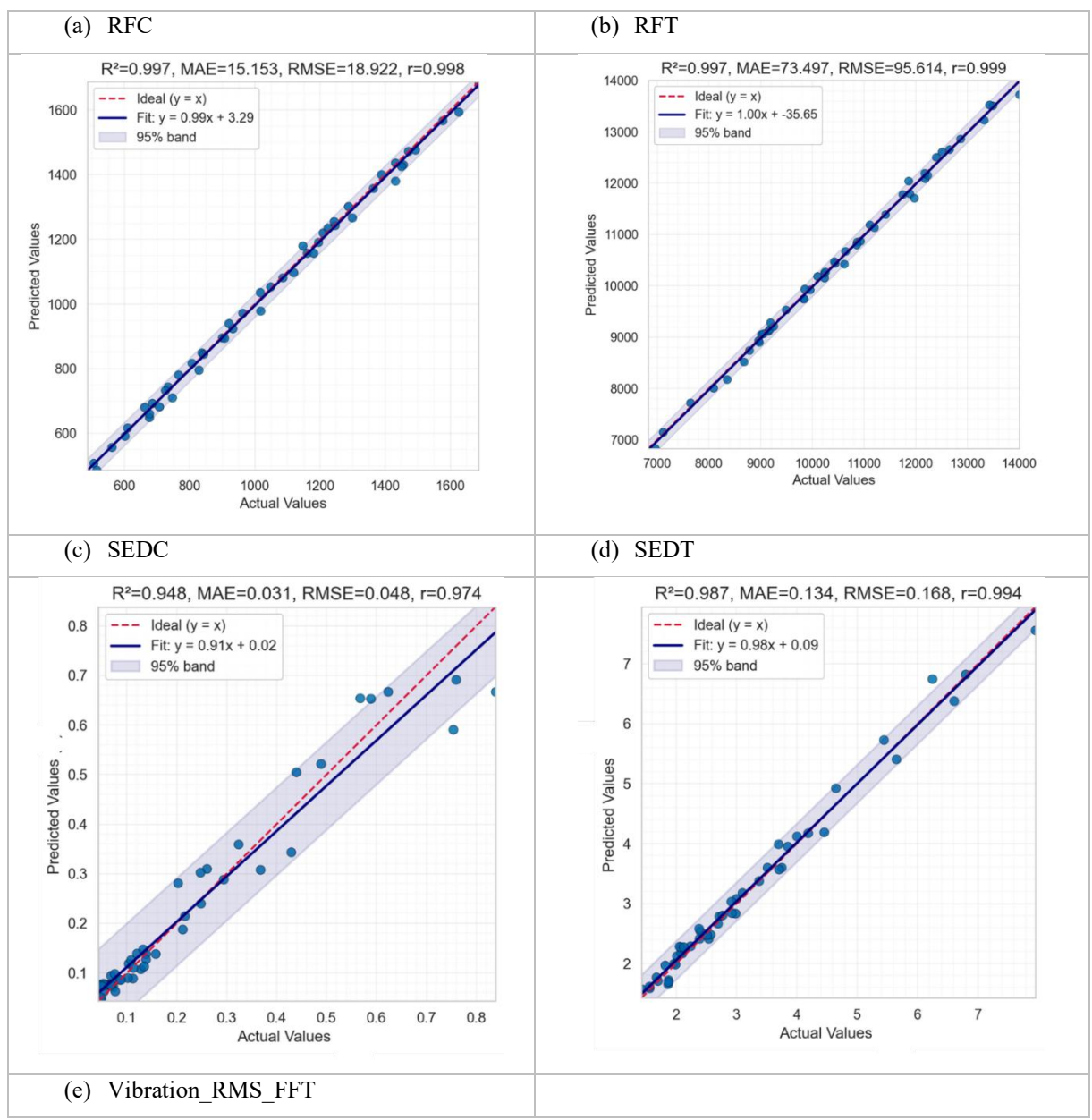

(a) RFC  (b) RFT  (c) SEDC  (d) SEDT  (e) Vibration_RMS_FFT

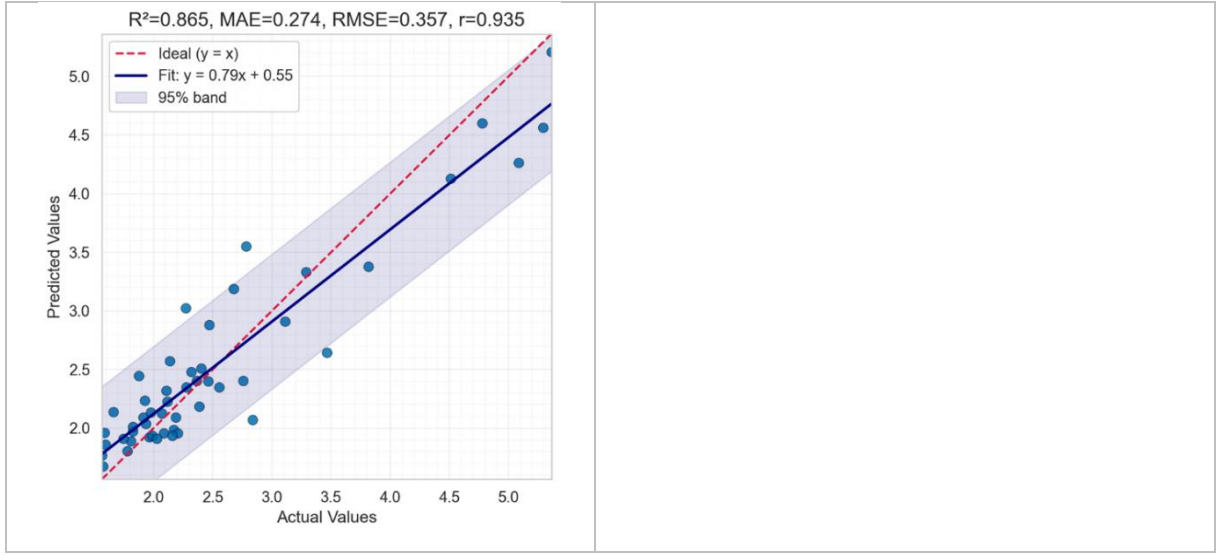

**Figure 15 Parity plots of output features (a) RFC (b) RFT (c) SEDC (d) SEDT and (e) Vibration_RMS_FFT with evaluation metrics for test set**

## 4.2 Single Objective Optimization of Stiffness, Durability and Vibration Behaviour:

Figures 16(a–f) collectively summarize the single-objective optimization results using PSO and BO for stiffness, durability, and vibration criteria. Figure 16(a) indicates that, for stiffness maximization and minimization (quantified by RFT), PSO reaches near-optimal values within the initial few iterations and then stabilizes, whereas BO exhibits larger oscillations before converging to a comparable region; notably, PSO achieves both high RFT (maximization) and low RFT (minimization) with faster and smoother convergence. The distribution of stiffness responses in Figure 16(b) corroborates this trend: the PSO-optimized designs occupy the extreme ends of the spectrum, well above and below the base design—demonstrating PSO's ability to efficiently traverse the design space to identify globally competitive solutions. For durability, where lower SEDT corresponds to higher durability, Figure 16(c) shows that PSO monotonically drives SEDT downward and stabilizes at a consistently lower level than BO, which fluctuates at higher SEDT values; this indicates that PSO is more effective at improving durability in this setting by persistently minimizing SEDT. The full-set distribution in Figure 16(d) further supports this observation: the optimized design is located near the lowest SEDT region (therefore highest durability) relative to the base design and most of the dataset, confirming successful navigation toward high-durability topologies. For vibration minimization, Figure 16(e) demonstrates that PSO again achieves a faster and more stable reduction in the RMS of the FFT vibration amplitude, whereas BO shows intermittent peaks before eventually approaching similar values; PSO's smoother trajectory suggests stronger exploitation and stability for this objective. Finally, Figure 16(f) shows the optimized

vibration-minimized design among the lowest responses across the dataset, markedly improving upon the base design. Taken together, these figures show that PSO consistently outperforms BO for stiffness maximization/minimization, durability enhancement (via SEDT minimization), and vibration minimization, achieving rapid, stable convergence to high-quality solutions across all single-objective cases.

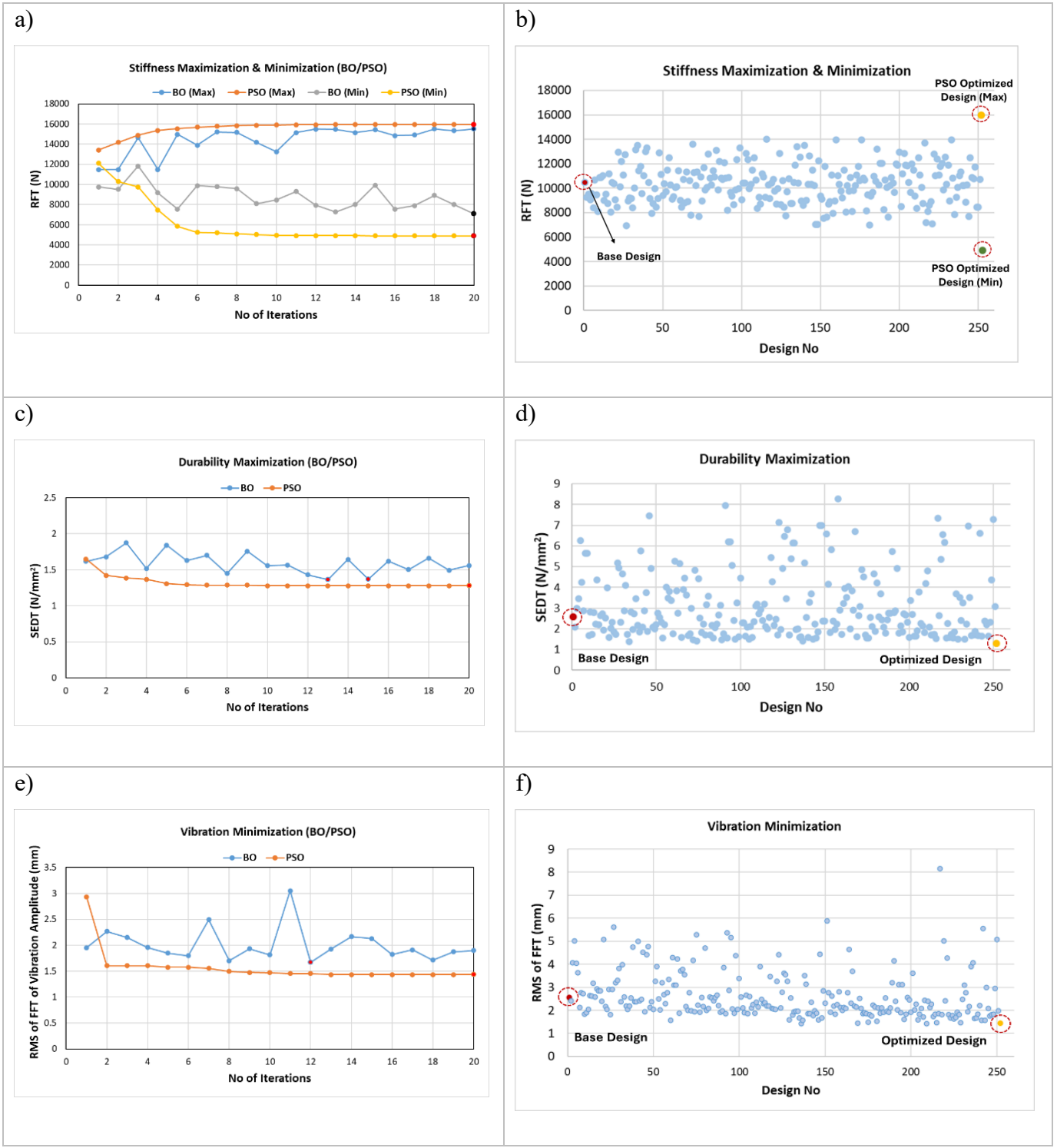

**Figure 16 PSO and BO convergence behaviour for (a, b) stiffness (max/min), (c, d) durability maximization, and (e, f) vibration minimization, shown in comparison with the initial dataset distribution and the base design**

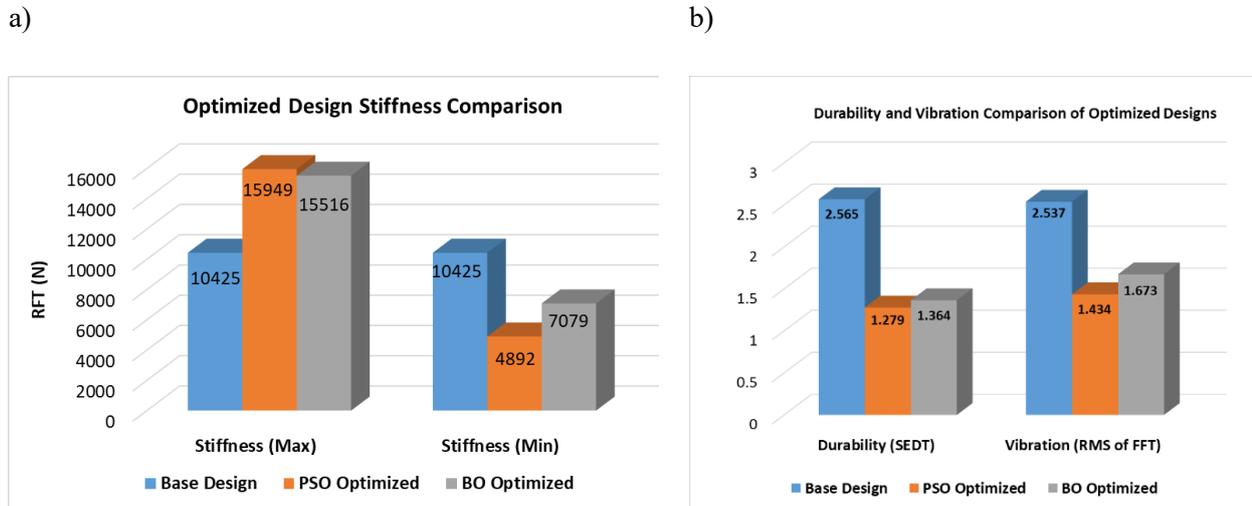

a)

b)

**Figure 17 Comparison of the PSO- and BO-optimized spoke designs with the base configuration in terms of a) stiffness b) durability and vibration characteristics.**

From Figure 17, it can be observed that the optimization framework enables substantial performance variation relative to the base spoke design across all targeted objectives. In the stiffness domain, the framework generates both highly stiff and highly compliant solutions: under stiffness maximization, RFT increases by **+52.99%** for PSO (Design B) and **+48.83%** for BO (Design C), while stiffness minimization yields RFT reductions of **−53.07%** (PSO) (Design D) and **−32.10%** (BO) (Design E). This wide tuning range demonstrates the capability of the spline-based parameterization to support both extremes of mechanical behavior. For durability, where **lower SEDT indicates higher durability**, the optimized designs show significant improvements, with SEDT reducing from **2.565 N/mm²** (base) to **1.279 N/mm²** (PSO) and **1.364 N/mm²** (BO), corresponding to **50.14%** (Design F) and **46.82%** (Design G) increases in durability, respectively. A similar trend appears in vibration performance, where the RMS amplitude decreases from **2.537 mm** (base) to **1.434 mm** (PSO) and **1.673 mm** (BO), giving **43.48%** (Design H) and **34.06%** (Design I) improvements. Collectively, these results indicate that while both optimizers outperform the baseline, PSO consistently produces the highest gains across stiffness tuning, durability enhancement, and vibration minimization. The design number A to I which is represented in the graph are generated and shown in Figure 18.

Complementing the quantitative improvements, Figure 18 visualizes how this performance gains relate to geometric changes in the spoke structure. All optimized designs were generated by applying the PSO- and BO-derived interpolation points to the top and bottom spline curves, enabling controlled shape evolution without altering the underlying parametric representation. The baseline configuration is shown in Figure 18 (A), while Figure 18 (B) to (I) illustrates representative optimized variations corresponding to key performance directions. These shapes reveal smooth yet meaningful adjustments in curvature, spline tension, and contour distribution, reflecting how targeted modifications in interpolation points translate into large mechanical effects such as increased stiffness, enhanced durability, or reduced vibration. Importantly, these visualizations confirm that the proposed spline-based generative approach can produce geometrically diverse and manufacturable spoke profiles spanning the full performance envelope—from significantly stiffer to substantially softer designs, as well as spokes optimized purely for durability or vibration resistance. Thus, the geometry set in Figure Y provides a direct qualitative link between parameter-level optimization and interpretable physical shape variation

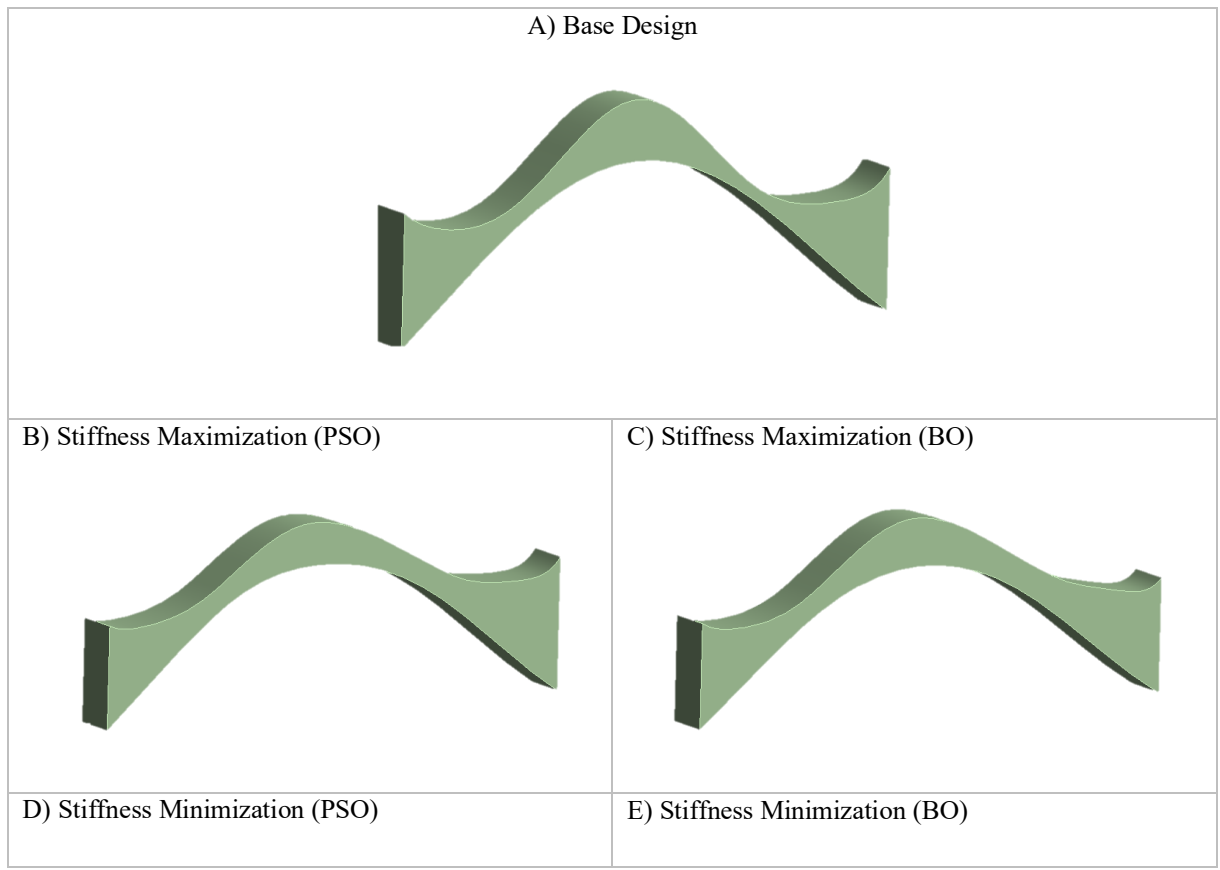

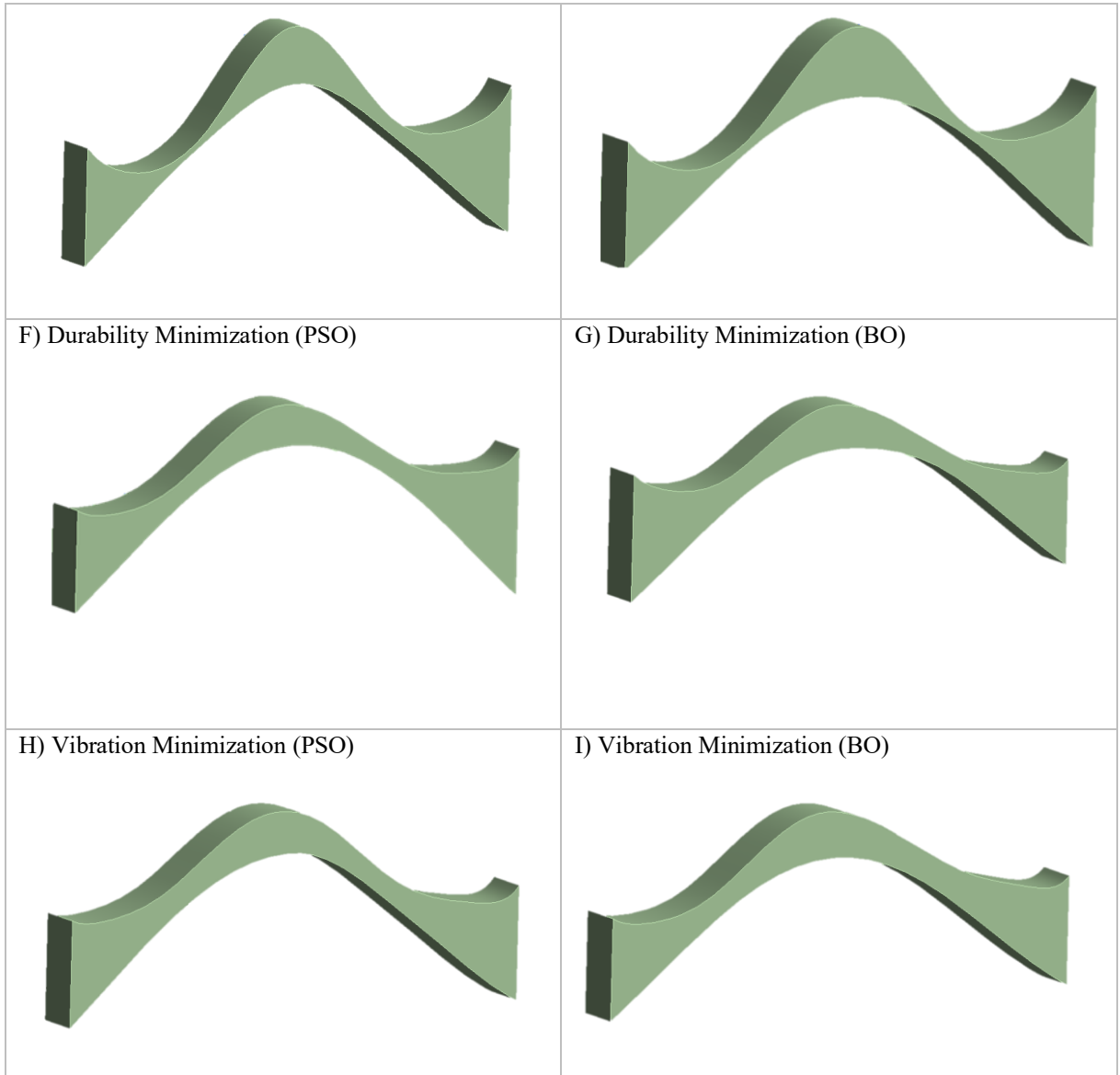

**Figure 18 (a) Base spoke design. PSO-generated spoke geometries for (b) maximum and (d) minimum stiffness, (f) durability minimization, and (h) vibration minimization. BO-generated spoke geometries for (c) maximum and (e) minimum stiffness, (g) durability minimization, and (i) vibration minimization**

### 4.3 Multi Objective Optimization of Stiffness, Durability and Vibration Behaviour:

#### *4.3.1 Min/Max Type Multi Objective Optimization:*

Figure 19(a) and Figure 19(b) illustrate the multi-objective optimization landscape for simultaneously maximizing stiffness (RFT), maximizing durability (minimizing SEDT), and minimizing vibration response. Using Bayesian Optimization (Figure 19 (a)), the design space is broadly explored, resulting in a widely scattered Pareto set that spans a large range of RFT (≈6000–14,000), SEDT (≈1–8), and vibration RMS values (≈1.8–4.5). The colour gradient shows that later BO iterations gradually move toward the region of lower vibration and lower SEDT, indicating progressive convergence but still maintaining significant variability due to the exploration-driven search behaviour. In contrast, the PSO-based multi-objective optimization (Figure 19(b)) produces a much more compact and coherent Pareto band, concentrated in the high-stiffness region (≈13,000–17,000 RFT) with consistently low SEDT (≈1.3–1.7) and reduced vibration RMS. The PSO iterations converge more tightly along a narrow trade-off surface, highlighting the algorithm's exploitative nature and its tendency to rapidly refine solutions around promising regions. Together, the two figures demonstrate that BO effectively maps the global structure of the design space, while PSO yields a sharper, high-quality Pareto front with superior stiffness–durability–vibration balance.

Figure 19(c) and Figure 19(d) show the multi-objective optimization results for the case where the optimizer seeks to minimize structural stiffness, maximize durability (i.e., minimize SEDT), and simultaneously minimize vibration response. In Figure 19(c), the Bayesian Optimization (BO) search exhibits broad coverage of the design space, with sampled solutions spanning low- to moderate-stiffness regions (≈6000–14,500 RFT), durability values in the range of ≈2–7 SEDT, and vibration levels between ≈2.0–4.5 RMS. The colour gradient shows that BO's early iterations explore widely separated configurations, while later iterations increasingly concentrate toward the desirable low-stiffness, low-SEDT, and low-vibration zone, though with considerable scatter due to BO's exploration-focused sampling. In contrast, the PSO-derived Pareto set in Figure 19(d) is much more compact and localized, with the swarm converging primarily to lower-stiffness solutions (≈7000–9000 RFT), reduced SEDT values (≈2–4), and consistently low vibration response (≈2.0–3.7 RMS). The tighter clustering of PSO solutions reflects its exploitative nature and its ability to rapidly converge toward high-quality trade-off regions. Overall, the two figures highlight that BO effectively maps the global space of feasible low-stiffness configurations, whereas PSO rapidly focuses on a narrow band of optimal low-stiffness, high-durability, and low-vibration designs.

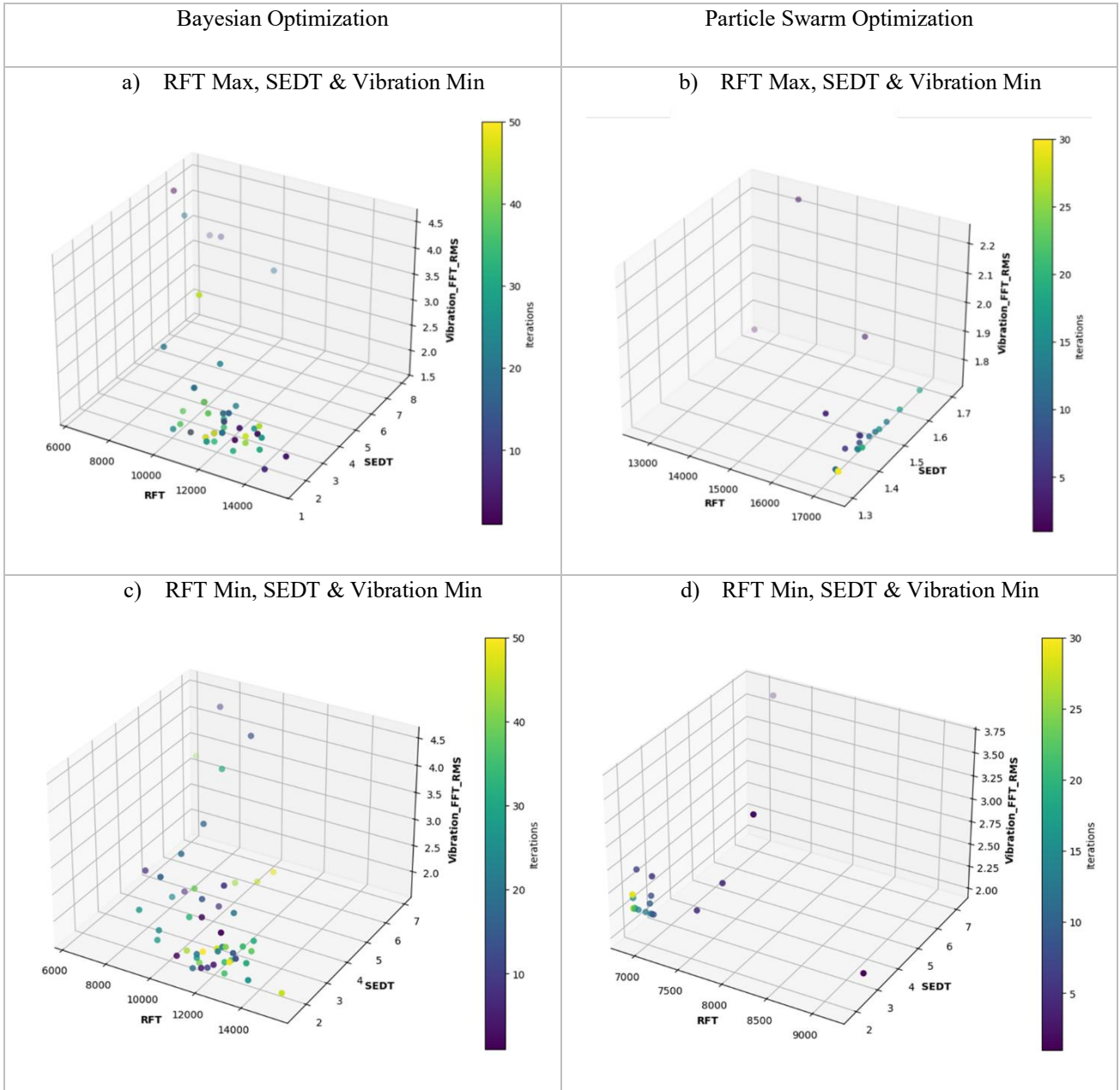

**Figure 19 Comparison of BO and PSO Pareto fronts for high-stiffness objectives (a–b) and low-stiffness objectives (c–d) with maximum durability and minimize vibration characteristics**

### 4.3.2 Targeted Multi Objective Optimization:

Figure 20(a–d) present the results of the targeted multi-objective optimization performed using Bayesian Optimization and PSO for achieving the prescribed performance goals of RFT ≈ 12,000, RFC ≈ 1000, maximum durability (i.e., minimum SEDT and SEDC), and minimum vibration response. Figures 20(a) and 20(b) correspond to the BO-driven targeted search. In Figure 20(a), which plots RFT–RFC–SEDT, BO exhibits broad exploration around and beyond the target region, with samples spread over RFT values from ≈6000–14,500, RFC values from ≈ 400–1800, and SEDT values between ≈ 1–8. This dispersion reflects BO's tendency to survey a wide range of feasible designs before progressively moving toward

low-SEDT configurations near the target stiffness and compliance. Figure 20(b), showing RFT–RFC–vibration, follows the same behaviour, with vibration amplitudes ranging from ≈2.0–5.0 RMS and later iterations (green, yellow) gradually migrating toward the region closer to the target RFC ≈ 1000 and reduced vibration levels. In contrast, Figures 20(c) and 20(d) illustrate the PSO-based targeted optimization, which displays a much more concentrated convergence pattern. PSO solutions cluster tightly around RFT ≈ 9500–12,500 and RFC ≈ 900–1150, with significantly reduced variability in both vibration (≈1.9–2.6 RMS in Figure 20(c)) and SEDC (≈0.07–0.13 in Figure 20(d)). This compact grouping indicates rapid exploitation of the target region, with swarm particles converging efficiently toward designs that simultaneously satisfy the stiffness, compliance, durability, and vibration constraints. Collectively, the four subfigures demonstrate that while BO provides broad global coverage and identifies several feasible performance regimes around the target, PSO achieves sharper convergence with higher precision around the desired multi-objective target point.

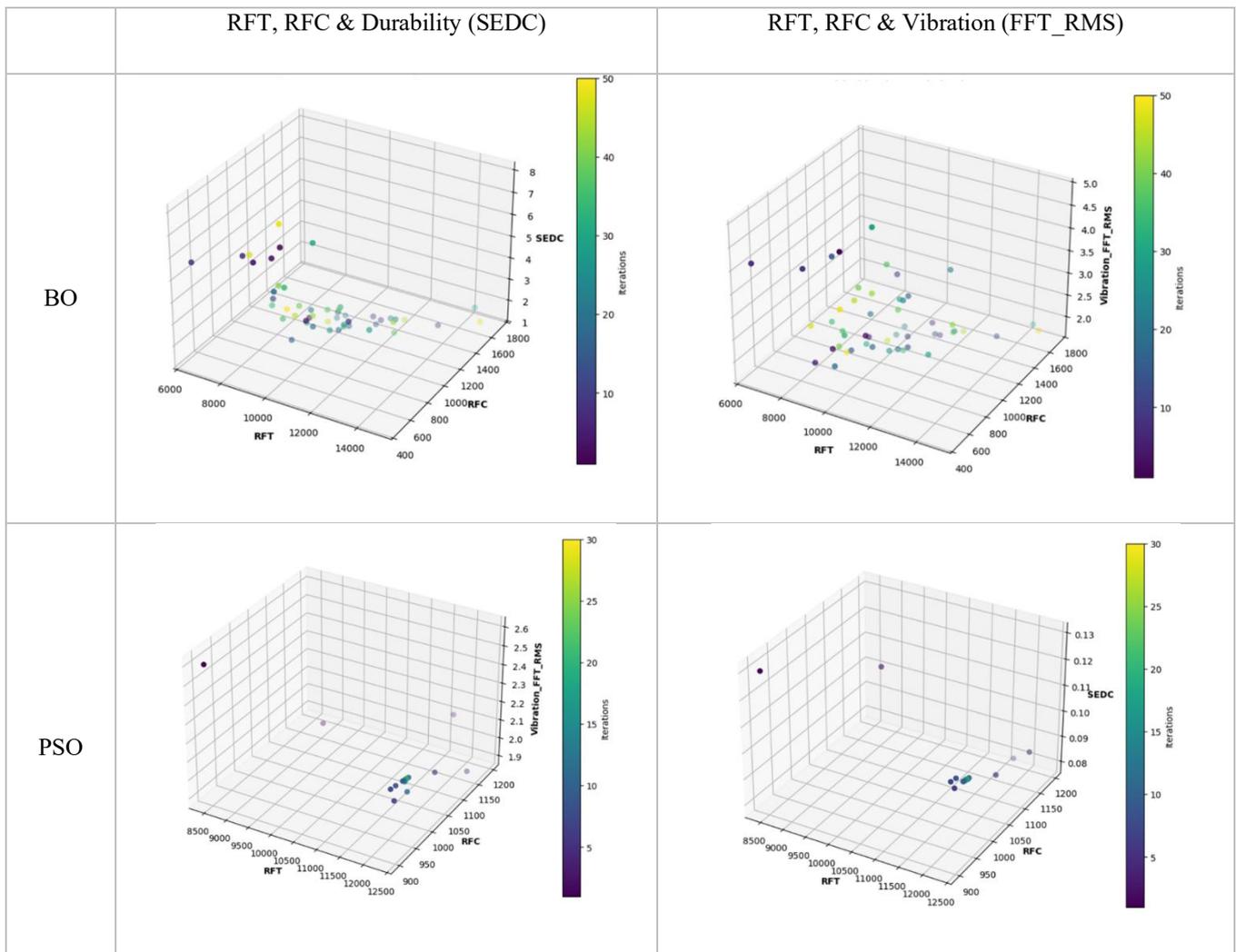

**Figure 20 Comparison of BO (a–b) and PSO (c–d) convergence behaviour for targeted stiffness, compliance, durability, and vibration objectives.**

# 5  Conclusion:

The generative design approach in combination with machine learning and FEM simulation is suggested for topological optimization of newly introduced NPT-UPTISs spoke design for passenger vehicle segment. It has been anlayzed that the of top and bottom curves of the spoke profiles have a remarkable impact on the stiffness, durability and vibration performance of whole NPT. The major findings of this study are as follows.

1) The top and bottom spoke profiles of the NPT-UPTIS structure can be accurately represented using 8th-order and 4th-order polynomial equations, respectively.
2) The **KRR model** efficiently captures stiffness behaviour (RFT and RFC) with $R^2$ values greater than 0.99, while **XGBoost** reliably predicts durability (SEDC and SEDT) with around 0.95 $R^2$ and vibration characteristics with approximately 0.86 $R^2$.
3) The proposed optimization framework enables the generation of spoke designs with **±53% stiffness tunability, up to ~50% improvement in durability**, and **up to ~43% reduction in vibration**, demonstrating its effectiveness in tailoring performance far beyond the base design.
4) **PSO** demonstrates **rapid and stable convergence**, making it highly effective for both **single-objective** and **targeted multi-objective optimization** where precise performance goals must be satisfied.
5) **Bayesian Optimization** offers superior **global exploration and is well-suited for multi-objective trade-off discovery** by generating diverse nondominated solutions and mapping the Pareto landscape.
6) This integrated, ML-guided optimization framework allows designing spoke structures with desired stiffness levels while simultaneously maximizing durability and minimizing vibration.